\documentclass[sigconf, nonacm]{acmart}

\usepackage{tikz}
\usepackage{bm}
\usepackage{amsmath,amsfonts}
 
\usepackage{amssymb}
\usepackage{pifont}
\usepackage{multirow}
\usepackage{hhline}
\usepackage{array,booktabs}
\usepackage{bbm}
\usepackage{wasysym}
\usepackage{pgfplots}
\usepackage{enumerate}
\usepackage{supertabular,booktabs}
\usepackage{ragged2e}
\usepackage{lineno}
\usepackage{wrapfig}
\usepackage[bottom]{footmisc}
\usetikzlibrary{pgfplots.groupplots}
\usetikzlibrary{patterns}
\usetikzlibrary{matrix}

\DeclareMathOperator*{\argmin}{arg\,min}

\newcommand{\cmark}{\ding{51}}

\definecolor{reb}{RGB}{0,0,0}

\newcommand\vldbdoi{10.14778/3461535.3461547}
\newcommand\vldbpages{XXX-XXX}
\newcommand\vldbvolume{14}
\newcommand\vldbissue{9}
\newcommand\vldbyear{2021}
\newcommand\vldbauthors{\authors}
\newcommand\vldbtitle{\shorttitle} 
\newcommand\vldbavailabilityurl{https://github.com/tedzhouhk/GCNP}

\begin{document}
\title{Accelerating Large Scale Real-Time GNN Inference using Channel Pruning}

\author{Hongkuan Zhou}
\author{Ajitesh Srivastava}
\author{Hanqing Zeng}
\affiliation{%
  \institution{University of Southern California}
  \city{Los Angeles}
  \country{USA}
}
\email{{hongkuaz,ajiteshs,zengh}@usc.edu}

\author{Rajgopal Kannan}
\affiliation{%
  \institution{US Army Research Lab}
  \city{Los Angeles}
  \country{USA}
}
\email{rajgopal.kannan.civ@mail.mil}

\author{Viktor Prasanna}
\affiliation{%
  \institution{University of Southern California}
  \city{Los Angeles}
  \country{USA}
}
\email{prasanna@usc.edu}

\begin{abstract}
  Graph Neural Networks (GNNs) are proven to be powerful models to generate node embedding for downstream applications. However, due to the high computation complexity of GNN inference, it is hard to deploy GNNs for large-scale or real-time applications. In this paper, we propose to accelerate GNN inference by pruning the dimensions in each layer with negligible accuracy loss. Our pruning framework uses a novel LASSO regression formulation for GNNs to identify feature dimensions (channels) that have high influence on the output activation. 
  We identify two inference scenarios and design pruning schemes based on their computation and memory usage for each.
  To further reduce the inference complexity, we effectively store and reuse hidden features of visited nodes, which significantly reduces the number of supporting nodes needed to compute the target embedding.
  We evaluate the proposed method with the node classification problem on five popular datasets and a real-time spam detection application. We demonstrate that the pruned GNN models greatly reduce computation and memory usage with little accuracy loss.
  For full inference, the proposed method achieves an average of $3.27\times$ speedup with only $0.002$ drop in F1-Micro on GPU. For batched inference, the proposed method achieves an average of $6.67\times$ speedup with only $0.003$ drop in F1-Micro on CPU.
  To the best of our knowledge, we are the first to accelerate large scale real-time GNN inference through channel pruning.
\end{abstract}

\maketitle

\begingroup\small\noindent\raggedright\textbf{PVLDB Reference Format:}\\
\vldbauthors. \vldbtitle. PVLDB, \vldbvolume(\vldbissue): \vldbpages, \vldbyear.\\
\href{https://doi.org/\vldbdoi}{doi:\vldbdoi}
\endgroup
\begingroup
\renewcommand\thefootnote{}\footnote{\noindent
This work is licensed under the Creative Commons BY-NC-ND 4.0 International License. Visit \url{https://creativecommons.org/licenses/by-nc-nd/4.0/} to view a copy of this license. For any use beyond those covered by this license, obtain permission by emailing \href{mailto:info@vldb.org}{info@vldb.org}. Copyright is held by the owner/author(s). Publication rights licensed to the VLDB Endowment. \\
\raggedright Proceedings of the VLDB Endowment, Vol. \vldbvolume, No. \vldbissue\ %
ISSN 2150-8097. \\
\href{https://doi.org/\vldbdoi}{doi:\vldbdoi} \\
}\addtocounter{footnote}{-1}\endgroup

\ifdefempty{\vldbavailabilityurl}{}{
\vspace{.3cm}
\begingroup\small\noindent\raggedright\textbf{PVLDB Availability Tag:}\\
The source code of this research paper has been made publicly available at \url{\vldbavailabilityurl}.
\endgroup
}

\section{Introduction}
\label{sec: intro}

Recently, Graph Neural Networks (GNNs) have attracted the attention of many AI researchers due to the high expressive power and generalizability of graphs in many applications. The node embedding generated from GNNs outperforms other graph representation learning methods when fed into downstream applications like node classification, edge prediction, and graph classification. Table \ref{tab: gnn application} shows some popular applications of GNNs on various size graphs with different latency requirements.
The knowledge graphs used in few-shot learning could only contain around one hundred of nodes and hundreds of edges, while the social network graphs could have billions of nodes and trillions of edges. 
Most of these GNN applications are latency sensitive at inference. For example, the applications related to Computer Vision need to perform streaming real-time inference on the data captured by the cameras. 
The applications related to fraud and spam detection need to identify malicious posts and transactions as fast as possible to avoid the property loss of the victim users.
In addition to latency, some vision applications that utilize GNNs \cite{qi20173d} need to perform inference on edge devices with limited computing power and memory, such as self-driving cars with 3D-cameras and radars.

\begin{figure}[b]
    \centering
\pgfplotsset{compat=newest}

\definecolor{ours}{RGB}{255,0,0}
\definecolor{gsage}{RGB}{0,0,153}
\definecolor{sgc}{RGB}{0,153,76}
\definecolor{sign}{RGB}{153,0,153}
\definecolor{pprgo}{RGB}{0,153,153}
\definecolor{gcn}{RGB}{64,64,64}
\definecolor{gat}{RGB}{153,153,0}
\definecolor{jk}{RGB}{102,0,204}
\definecolor{tinygnn}{RGB}{255,128,0}

\includegraphics{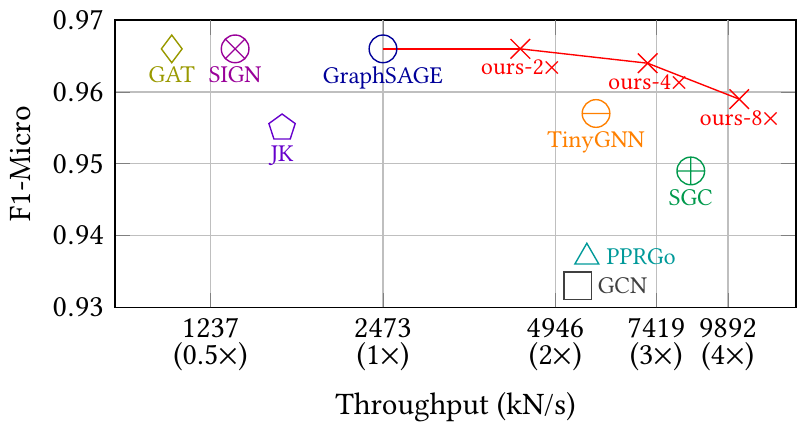}
    \caption{Accuracy and throughput of full inference on the Reddit dataset on GPU.}
    \label{fig: acct}
\end{figure}

Compared with traditional graph analytics algorithms, GNNs have high computation cost as one node needs to gather and aggregate feature vectors from all the neighbors in its receptive field to compute a forward pass. To accelerate the training of GNNs, many works \cite{graphsage,graphsaint,fastgcn,s-gcn} adopt stochastic node sampling techniques to reduce the number of supporting neighbors.
\textcolor{reb}{GraphNorm \cite{cai2020graphnorm} normalizes the node attributes to speedup the convergence.}
As a result, GNNs training scales well with graph size. It only takes seconds to minutes to train on a graph with millions of nodes.
However, many GNN applications struggle at inference when deployed to production environment.
Performing the full forward pass with all the neighbors at inference leads to high memory usage and latency. 
The node sampling techniques, when applied to inference, struggle to maintain high accuracy on every sample. 
In consequence, GNN applications either turn to traditional graph analytics algorithms with lower complexity, or rely on obsolete (not updated recently) embedding. For example, Youtube \cite{halcrow2020grale} turns to label propagation to detect abusive videos. Pinterest \cite{pinterest} has to use obsolete embedding generated with the MapReduce framework in an offline process. Taobao \cite{liuheterogeneous} runs the GNN based malicious account detection daily, instead of immediately after one transaction pops. Even with the compromise of offline inference, GNN inference is still expensive on large graphs. It is reported that a cluster with 378 computing nodes still needs one day to generate embedding for 3 billion nodes \cite{pinterest}. 
In addition, GraphBERT \cite{zhang2020graph} shows that pre-trained GNN models could be directly (or with light fine-tuning) transferred to address new tasks, which makes accelerating GNN inference more important.

\begin{table}[t!]
    \centering
    \caption{\textcolor{reb}{GNN Applications with their conventional graph sizes (in number of nodes) and latency requirement.}}
    \begin{tabular}{>{\centering\arraybackslash}m{2.0in}|c|c}
        Applications & Nodes & Lat.\\
        \toprule
        \multicolumn{3}{c}{\textbf{Knowledge Graph}}\\
        \midrule
        Few-shot image classification \cite{garcia2018fewshot,gidaris2019generating} & \textcolor{reb}{$10^2-10^3$} & ms \\
        \midrule
        Relation extraction and reasoning \cite{schlichtkrull2018modeling} & \textcolor{reb}{$10^3-10^6$} & ms-s \\
        \midrule
        \multicolumn{3}{c}{\textbf{Image Graph}}\\
        \midrule
        Point cloud segmentation \cite{wang2019dynamic,qi2017pointnet} & \textcolor{reb}{$10^3-10^6$} & ms\\
        \midrule
        \multicolumn{3}{c}{\textbf{Spatio-Temporal Graph}}\\
        \midrule
        Traffic prediction \cite{guo2019attention} & \textcolor{reb}{$10^3-10^6$} & s\\
        \midrule
        Action recognition \cite{yan2018spatial,qi2018learning} & \textcolor{reb}{$10^2-10^3$} & ms\\
        \midrule
        \multicolumn{3}{c}{\textbf{Social Network Graph}}\\
        \midrule
        Recommending system \cite{pinterest,zhang2019star,fanmetapath} & \textcolor{reb}{$10^6-10^9$} & ms\\
        \midrule
        Spam detection \cite{xianyu,wang2019semi,liuheterogeneous} & \textcolor{reb}{$10^6-10^9$} & ms\\
    \end{tabular}
    \label{tab: gnn application}
\end{table}

Although it has not caught much attention of researchers, accelerating GNN inference is as important as accelerating GNN training. Based on these GNN applications, we define two inference scenarios -- full inference where the target nodes are all the nodes (or a large portion of nodes, i.e., the test set) in the graph, and batched inference where the target nodes are a few nodes. Full inference applies to GNN applications that operate on small to medium size graphs, or perform offline inference on large graphs. Batched inference applies to GNN applications that have strict requirements on latency, or need to be executed on edge devices such as embedded systems and FPGAs. Full inference performs forward propagation on all the nodes in the graph, while batched inference only propagates from the selected supporting nodes of the target nodes. For batched inference, the number of supporting nodes grows exponentially with the number of GNN layers, which is referred to as the ``neighbor explosion'' problem. 
In this work, we propose to accelerate GNN inference by reducing the input feature dimensions in each GNN layer and reusing the hidden features for visited nodes. Our pruning framework works on most GNN architectures and can significantly improve their inference throughput with little or no loss in accuracy. 
The main contributions of this work are

\begin{itemize}
    \item We develop a novel LASSO regression formulation to prune input channels for GNN layers, which outperforms random and greedy pruning methods.
    \item We design different pruning schemes for full inference and batched inference addressing their computation complexity and memory usage.
    \item We develop a novel technique to store and reuse the hidden features of visited nodes for batched inference, which mitigates the ``neighbor explosion'' problem.
    \item We evaluate the performance of the pruned models on five popular datasets \textcolor{reb}{and a real-time spam detection application}. The pruned GNN models greatly reduce the complexity and memory usage with negligible accuracy loss. For full/batched inference, the pruned models reduce the computation to $0.19\times$/$0.10\times$ and memory requirements to $0.43\times$/$0.18\times$ with only $0.002$/$0.003$ F1-Micro drop on average. The pruned models achieve an average of $3.27\times$/$6.67\times$ speedup for full/batched inference on GPU/CPU.
\end{itemize}

\section{Background}

\begin{figure*}
    \centering
    \resizebox{0.8\textwidth}{!}
    {
\includegraphics{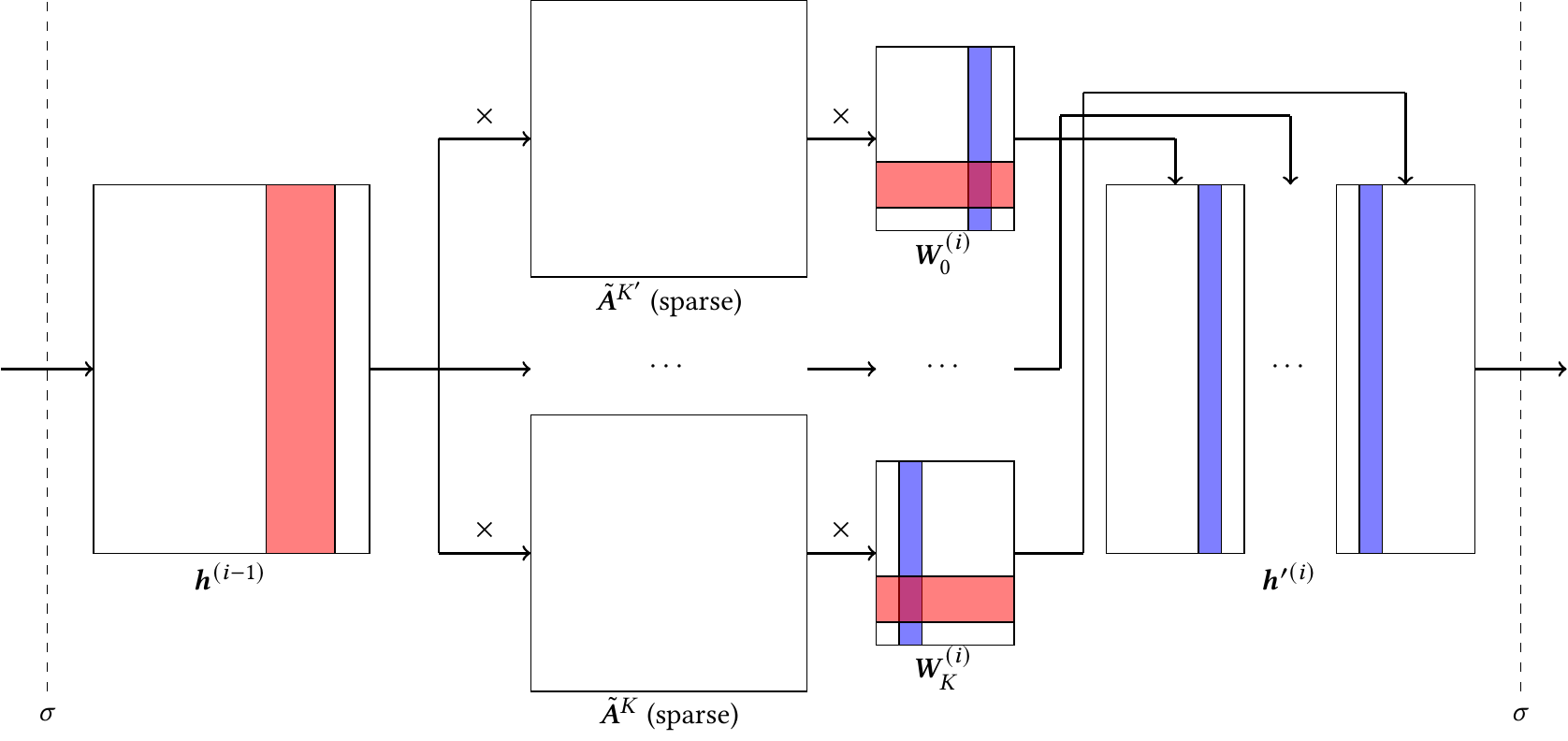}
    }
    \caption{Illustration of one pruned GNN layer. The $\times$ operator denotes sparse or dense matrix multiplication while the shaded areas denote the pruned channels. The blue areas in the weight matrices $\bm{W}_k^{(i)}$ and the output features $\bm{h}'^{(i)}$ before activation show the pruned channels in the next GNN layer. The red areas in weight matrices $\bm{W}_k^{(i)}$ and the input features $\bm{h}^{(i-1)}$ show the pruned channels in this GNN layer.}
    \label{fig:prune}
\end{figure*}

\subsection{Graph Neural Networks}

For a graph $G(V,E)$ where each node $v\in V$ has node attributes $\bm{h}(v)\in\mathbb{R}^f$, GNNs iteratively gather and aggregate information from neighbors to compute node embedding. Denote the matrix of all the output features $\bm{h}^{(i)}(v)$ stacked horizontally in layer-$i$ by $\bm{h}^{(i)}$. Let $\bm{\tilde{A}}$ be the normalized adjacency matrix. In general, the output features $\bm{h}^{(i)}$ of layer-$i$ is computed by

\begin{equation}
    \bm{h}^{(i)}=\sigma\left(\mathop{\Vert}_{k=K'}^K\bm{\tilde{A}}^k\bm{h}^{(i-1)}\bm{W}_k^{(i)}\right)
    \label{eq: forward}
\end{equation}
where $\Vert$ denotes the horizontal concatenation operation. $\bm{W}_k^{(i)}$ is the learnable weight matrix of order $k$ in layer-$i$. And $\sigma(\cdot)$ denotes the ReLU activation. We stack multiple layers and let the input of the first layer $\bm{h}^{(0)}=\bm{h}$ to compute the node embedding. For $K'=K=1$, Equation \ref{eq: forward} shows the forward propagation of vanilla Graph Convolutional Network \cite{kipfgcn}. For $K'=0,K=1$, Equation \ref{eq: forward} is the GraphSAGE \cite{graphsage} architecture. For $K'=0,K>1$, Equation \ref{eq: forward} is the MixHop \cite{mixhop} architecture. 
For other variants of GNNs \cite{gat,jumpingknowledge,gin}, Equation \ref{eq: forward} could be adapted by adding residue connections or alternating the normalized adjacency matrix.

\subsection{Case Study: GraphSAGE Inference}
\label{sec: case}

We perform a case study to analyze the complexity and memory usage for both inference scenarios on the widely used GraphSAGE architecture. We choose to analyze the GraphSAGE architecture as it achieves top tier accuracy with relatively high throughput (see Figure \ref{fig: acct}). For the GraphSAGE architecture, $K'=0,K=1$ and the adjacency matrix $\bm{A}$ is normalized by $\tilde{\bm{A}}=\bm{D}^{-1}\bm{A}$ where $\bm{D}$ is the diagonal degree matrix.

\subsubsection{Full Inference}

To perform full inference that computes node embedding for all the nodes in the graph, we \textcolor{reb}{batch the node-wise aggregation and compute sparse-dense matrix multiplication $\bm{\tilde{A}}\cdot\bm{h}^{(i-1)}$.}
Denote the input and output feature dimensions of the weight matrices $\bm{W}_{k}^{(i)}$ by $f_{k}^{\text{in}(i)}$ and $f_{k}^{\text{out}(i)}$ (all input feature dimensions are equal in each layer). Let $|V|$ be the number of nodes in the graph. Assume the average degree of the whole graph is $d$. The average complexity per node $C^{(i)}_{\text{full}}$ and total memory consumption $M^{(i)}_{\text{full}}$ of full inference are

\begin{equation}
    \begin{aligned}
        C^{(i)}_{\text{full}}&=\mathcal{O}\left(d\min (f_{1}^{\text{in}(i)},f_1^{\text{out}(i)})+\sum_{k=0}^1f_{k}^{\text{in}(i)}f_{k}^{\text{out}(i)}\right)\\
        M^{(i)}_{\text{full}}&=|V|\left(f_{0}^{\text{in}(i)}+f_{0}^{\text{out}(i)}+f_{1}^{\text{out}(i)}\right)+\sum_{k=0}^1f_{k}^{\text{in}(i)}f_{k}^{\text{out}(i)}\\
    \end{aligned}
\end{equation}
As the output features of all nodes are computed in every layer, the computation and memory consumption distribute evenly in each layer. \textcolor{reb}{Each branch in one layer also contributes to a non-negligible portion of the computation and memory usage.} 

\subsubsection{Batched Inference}

For batched inference, the GraphSAGE architecture aggregates from $L$-hop neighbors. Denote the set of target nodes to infer by $V_t$. In layer-$i$, the average number of supporting nodes is $|V_t|\sum_{l=0}^{L-i+1}d^l$,
\textcolor{reb}{which leads to the average complexity per node dominated by the complexity in the last layer}

\begin{equation}
    C_{\text{batched}}=\sum_{i=1}^LC_{\text{batched}}^{(i)}=\sum_{i=1}^L\sum_{l=0}^{L-i}d^lC_{\text{full}}^{(i)}=\mathcal{O}(d^{L-1}C_{\text{full}}^{(1)})
    \label{eq: c_batched}
\end{equation}
\textcolor{reb}{Similarly, the memory consumption also peaks in the first layer with the most supporting neighbors.}

\subsection{Related Work}
\label{sec: relatedwork}

There are many existing works on channel pruning in Deep Neural Networks. 
\textcolor{reb}{
The works \cite{he2017channel,10.5555/3157096.3157329} prune the channels in the convolution layer by applying penalized regression on the input channels.
ThiNet \cite{thinet} prunes the channels based on statistics from the next layer.
The work \cite{ye2018rethinking} forces some channels to freeze during the training and remove them at inference.
}
\textcolor{reb}{Unlike performing inference on texts or images where each instance is independent with the others, inference of nodes depends on the graph structure and attributes of other supporting nodes. The computation pattern is also different for different inference scenarios. These two challenges make it hard to directly apply the existing channel pruning techniques on GNNs.}
Recently, several works \cite{sgc,frasca2020sign} have tried to accelerate training and inference of GNNs by removing the nonlinearity of internal layers and pre-computing the feature aggregation ($\bm{A}^k\bm{h}$).
PPRGo \cite{PPRGo} accelerates inference by performing less aggregation as in training. These methods require pre-processing on either the node attributes or the adjacency matrix, which do not apply to evolving graphs.
TinyGNN \cite{tinygnn} speeds up inference by training a shallow student GNN supervised by a teacher GNN. 
\textcolor{reb}{Recently, several works have tried to accelerate the full batch propagation in Equation \ref{eq: forward} through matrix partitioning \cite{graphsaint-ipdps19}, node re-ordering \cite{fullinffpga}, and runtime scheduling \cite{9139807}. Others have developed hardware accelerators \cite{9065592,Zeng_2020} and in-memory processors \cite{8327035}. These hardware-specific optimization techniques do not address the basic problem -- high computation complexity of GNNs.}
On the other hand, although not aiming at rapid inference, some works \cite{zhang2018graph,Xu2020Dynamically} propose to prune the edges to reduce the noise aggregated from neighbors. However, they are limited to knowledge graphs with specific inference queries.

In contrast, we propose a general method to reduce the inference complexity by directly pruning the input channels.
\textcolor{reb}{
Our method works for all types of graphs and most GNN architectures. Combined with edge pruning methods and architecture simplification methods, our method has the potential to further speed up the inference.     
In addition, our pruning method does not incur extra sparse operations. The dimensions of the matrix operations are also lower, which makes it easier to design hardware accelerators and in-memory processors.
}
 
\section{Approach}

In GNNs, channels refer to column vectors in the hidden features matrix $\bm{h}^{(i)}$. We propose to solve the channel pruning problem by applying LASSO regression \cite{tibshirani1996regression} directly on the input channels. %
For a pre-trained GNN model, we prune the channels reversely from the output layer to the input layer. 
Figure \ref{fig:prune} shows one pruned GNN layer with multiple branches.
In this section, we first introduce the formulation and optimization of channel pruning in a single layer. Then, we discuss the pruning schemes for full inference and batched inference.
For batched inference, we further propose a novel technique that stores the hidden features for visited nodes and aggregates directly from them during inference. 

\subsection{Single Branch Pruning}

To prune the channels, we aim to generate the same output features in a branch before activation $\bm{h}'^{(i)}$ with fewer input dimensions. We focus on $\bm{h}'^{(i)}$ instead of $\bm{h}^{(i)}$ to keep all operations linear. For branch $k$ ($K'\leq k\leq K$) in layer-$i$, let $c_k^{(i)}=f_k^{\text{in}(i)}$ be the number of channels in the original GNN. We formulate the channel pruning problem with budget $\eta_k^{(i)}$ as the following optimization problem

\begin{equation}
    \begin{aligned}
        &\argmin_{\hat{\bm\beta}_k^{(i)},\widehat{\bm{W}}_k^{(i)}}\left\|\bm{Y}_k^{(i)}-\tilde{\bm{A}}^k\bm{h}^{(i-1)}\odot\hat{\bm\beta}_k^{(i)}\widehat{\bm{W}}_k^{(i)}\right\|_2^2\\
        &\text{subject to }\left\|\hat{\bm\beta}_k^{(i)}\right\|_0\leq\eta_k^{(i)}c_k^{(i)}
    \end{aligned}
\end{equation}
where $\bm{Y}_k^{(i)}=\tilde{\bm{A}}^k\bm{h}^{(i-1)}\bm{W}_k^{(i)}$ is the target output features. $\left\|\cdot\right\|_2$ is the L2-norm and $\left\|\cdot\right\|_0$ is the L0-norm measuring the number of non-zero elements. $\hat{\bm\beta}_k^{(i)}\in\mathbb{R}^{c_k^{(i)}}$ is the coefficient vector acting as masks for each channel. $\odot$ denotes element-wise multiplication on each row of the matrix. If $\hat{\bm\beta}_k^{(i)}(j)=0$, then the $j^\text{th}$ channel of the input features $\bm{h}^{(i-1)}$ can be removed. To solve the optimization problem, we first relax the L0-norm to L1-norm and add a penalty term with penalty factor $\lambda$.

\begin{equation}
    \begin{aligned}
        &\argmin_{\hat{\bm\beta}_k^{(i)},\widehat{\bm{W}}_k^{(i)}}\left\|\bm{Y}_k^{(i)}-\tilde{\bm{A}}^k\bm{h}^{(i-1)}\odot\hat{\bm\beta}_k^{(i)}\widehat{\bm{W}}_k^{(i)}\right\|_2^2+\lambda\left\|\hat{\bm\beta}_k^{(i)}\right\|_1\\
    \end{aligned}
\end{equation}

We separate the optimization of $\hat{\bm\beta}_k^{(i)}$ and $\widehat{\bm{W}}_k^{(i)}$ into two sub-problems and optimize on the sub-problems iteratively to find the global minimum. Initially, $\widehat{\bm{W}}_k^{(i)}=\bm{W}_k^{(i)}$ and $\hat{\bm\beta}_k^{(i)}=\mathbbm{1}$. We optimize both sub-problems on the hidden features of the training nodes. We use the training graph \textcolor{reb}{as the normalized adjacency matrix during optimization to avoid information leak}.

\subsubsection{Optimization on $\hat{\bm\beta}_k^{(i)}$}

To optimize $\hat{\bm\beta}_k^{(i)}$, $\widehat{\bm{W}}_k^{(i)}$ is fixed. The problem becomes a classic LASSO regression problem with ``large $n$, small $p$''. We solve the LASSO regression with Stochastic Gradient Descent (SGD). Due to the L1-norm term in the constraint, some mask values in the solution of $\hat{\bm\beta}_k^{(i)}$ would shrink to zero, leading to the removal of the corresponding channels. 

\begin{equation}
    \begin{aligned}
        &\argmin_{\hat{\bm\beta}_k^{(i)}}\left\|\bm{Y}_k^{(i)}-\bm{Z}_k^{(i)}\odot\hat{\bm\beta}_k^{(i)}\right\|_2^2+\lambda\left\|\hat{\bm\beta}_k^{(i)}\right\|_1\\
    \end{aligned}
\end{equation}
where $\bm{Z}_k^{(i)}=\tilde{\bm{A}}^k\bm{h}^{(i-1)}\widehat{\bm{W}}_k^{(i)}$.

\subsubsection{Optimization on $\widehat{\bm{W}}_k^{(i)}$} To optimize $\widehat{\bm{W}}_k^{(i)}$, $\hat{\bm\beta}_k^{(i)}$ is fixed. The problem becomes a quadratic programming problem

\begin{equation}
    \argmin_{\widehat{\bm{W}}_k^{(i)}}\left\|\bm{Y}_k^{(i)}-\bm{X}_k^{(i)}\widehat{\bm{W}}_k^{(i)}\right\|_2^2
\end{equation}
where $\bm{X}_k^{(i)}=\tilde{\bm{A}}^k\bm{h}^{(i-1)}\odot\hat{\bm\beta}_k^{(i)}$. The closed-form least square solution is given by $\widehat{\bm{W}}_k^{(i)}=(\bm{X}_k^{(i)\top}\bm{X}_k^{(i)})^{-1}\bm{X}_k^{(i)\top}\bm{Y}_k^{(i)}$.

\subsection{Single Layer Pruning}
\label{sec: single layer pruning}

To prune the input channels for one layer with multiple branches, we need to ensure each branch shares the same pruned channels in $\bm{h}^{(i-1)}$ so that these channels could be removed in the output of the previous layer. We enforce that the same channels are pruned in each branch by applying a shared coefficient vector $\bm{\hat{\beta}}^{(i)}$ and jointly optimizing on all the branches.

\begin{equation}
    \begin{aligned}
        &\argmin_{\hat{\bm\beta}^{(i)},\widehat{\bm{W}}_k^{(i)}}\sum_{k=K'}^{K}\left\|\bm{Y}_k^{(i)}-\tilde{\bm{A}}^k\bm{h}^{(i-1)}\odot\hat{\bm\beta}^{(i)}\widehat{\bm{W}}_k^{(i)}\right\|_2^2+\lambda\left\|\hat{\bm\beta}^{(i)}\right\|_1\\
    \end{aligned}
\end{equation}
The sub-problem of $\bm{\hat{\beta}}^{(i)}$ forms a generalized LASSO optimization problem.

\begin{equation}
    \begin{aligned}
        &\argmin_{\hat{\bm\beta}^{(i)}}\left\|\bm{Y}^{(i)}-g\left(\bm{h}^{(i-1)}\odot\hat{\bm\beta}^{(i)}\right)\right\|_2^2+\lambda\left\|\hat{\bm\beta}^{(i)}\right\|_1\\
    \end{aligned}
\end{equation}
where $\bm{Y}^{(i)}=\Vert_{k=K'}^K\bm{Y}_k^{(i)}$ stacked horizontally. And the generalized function $g(\bm{h}'^{(i-1)})=\Vert_{k=K'}^{K}\tilde{\bm{A}}^k\bm{h}'^{(i-1)}\widehat{\bm{W}}_k^{(i)}$ stacked horizontally. However, as $\bm{Y}^{(i)}$ is also concatenated horizontally, we can rewrite the LASSO optimization problem by substituting the original observations $\bm{h}^{(i-1)}$ with $\tilde{\bm{A}}^k\bm{h}^{(i-1)}$ concatenated vertically. Then, the sub-problem of $\bm{\hat{\beta}}^{(i)}$ falls back to a classic LASSO regression with $(K-K')$ times the observations in the single branch pruning and the same number of predictors.

\textcolor{reb}{Our pruning method also works for other GNN architectures. For GNN architectures with averaging instead of concatenating the output features in each branch, the generalized function becomes $g(\bm{h}^{(i-1)})=\sum_{k=K'}^K\tilde{\bm{A}}^k\bm{h}^{(i-1)}\widehat{\bm{W}}_k^{(i)}/(K-K'+1)$, which can be optimized by concatenating the observations horizontally. For multi-head attention based architecture, we can prune the layers by treating each attention head as a branch.}

\begin{figure}[t]
    \centering
\includegraphics{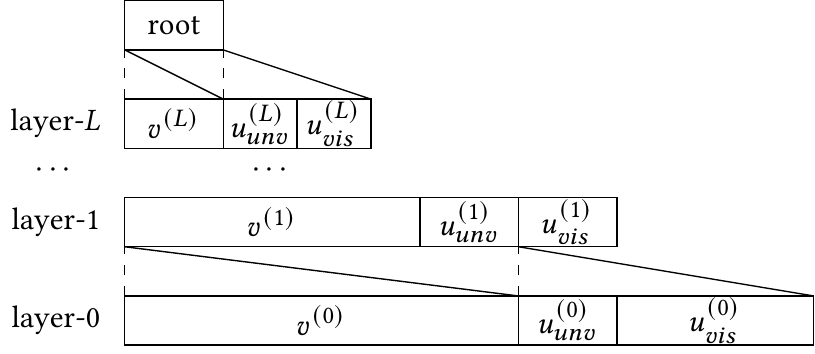}

    \caption{Illustration of forward propagation on $L+1$-layer GraphSAGE architecture with stored hidden features. $v^{(i)}$ denotes the supporting nodes for the branch $k=0$. $u_{unv}^{(i)},u_{vis}^{(i)}$ denote the unvisited and visited supporting nodes for the branch $k=1$. The hidden features of the visited nodes $u_{vis}^{(i)}$ are obtained directly from the stored hidden features.}
    \label{fig: batched}
\end{figure}

\subsection{End-to-End GNN Pruning}

We design the pruning schemes for different inference scenarios based on the computation complexity and memory consumption in the case study in Section \ref{sec: case}. The channel pruning in layer-$i$ not only ignores some input channels in $\bm{h}^{(i-1)}$, but also leads to the reduction of output channels in the weight matrices $\bm{W}_k^{(i)}$ of the previous layer. Thus, when pruning the whole network, we prune reversely from the output layer to the input layer. Dense layers are treated as GNN layers with $K'=K=0$.

\subsubsection{Pruned Full Inference}
\label{sec: prunedfullinf}

For full inference, we simply prune each layer with a constant budget $\eta$ except the input layer as the computation and memory distribute evenly in each layer. We do not remove any dimensions of the raw node attributes in layer-$0$. 
\textcolor{reb}{
The complexity per node and memory usage of the pruned models range in $(\eta^2,\eta)$ and $(\eta,1)$, compared with the original model.
For other GNN architectures that follow similar forward propagation in Equation \ref{eq: forward} like JK \cite{jumpingknowledge} and SIGN \cite{frasca2020sign}, our pruning method could be directly applied with constant budget.}

\subsubsection{Pruned Batched Inference}
\label{sec: prunedbatchinf}

The major challenge in batched inference is the ``neighbor explosion'' problem where the number of supporting nodes grows exponentially as the network goes deeper. We need to visit the node attributes for an exponential amount of nodes to compute the embedding for one target node. Therefore, we focus on reducing the computation and memory usage \textcolor{reb}{in the first layer by reducing the channels in the first layer and the second layer. In the first layer, we focus on the branches that have more neighbors than others.} For the GraphSAGE architecture with pruning budget $\eta$, we prune the $k=1$ branch in layer-$1$ and the whole layer-$2$ with budget $\eta$, which reduces the dominant terms in the computation and memory usage by $\eta$. 

In addition to channel pruning, we store the hidden features $\bm{h}^{(i)}$ of visited nodes in the middle layers. Their neighbors, when aggregating from them, directly aggregate from the stored hidden features, instead of iteratively looking at farther neighbors. Figure \ref{fig: batched} shows the supporting nodes in each layer with stored hidden features. Ideally, if we store the hidden features for all visited nodes, the batched inference would have exactly the same complexity as full inference (i.e., $d=1$ in Equation \ref{eq: c_batched}). However, indexing and storing the hidden features incur extra data transfer which increases the latency. 
On evolving graphs, out-dated hidden features also affect accuracy.
\textcolor{reb}{The portion of hidden features to store in each batch could be dynamically determined by the task-specific target latency and accuracy. Applications with high latency tolerance could potentially save more hidden features to increase throughput. For out-dated hidden features, we can set a threshold and discard them when the accuracy drop reaches the threshold. In practice, we find storing the hidden features for the root nodes at inference is a good balance point for the datasets we use. In addition, the root nodes usually have the most up-to-date hidden features in batched inference.}

\subsubsection{Detailed Optimization Procedure}
\label{sec: detailopt}

In the experiment, we perform one iteration on each sub-problem instead of multiple iterations \cite{he2017channel}. For the sub-problem of $\widehat{\bm{W}}$, instead of the least square solution, we also apply SGD as the size of $\bm{X}$ could be large. We partition the matrix $\tilde{\bm{A}}^k\bm{h}^{(i-1)}$ and $\bm{X}_k^{(i)}$ row-wise to form mini-batches. 
Define one epoch as performing SGD on the whole matrix once. To optimize the whole problem, we first optimize several epochs on the sub-problem of $\hat{\bm\beta}$. At the end of each epoch, we slightly increase the penalty factor $\lambda$ until \textcolor{reb}{pruning budget is met or over-penalized (all values in $\bm\beta$ are decreasing)}. Note that as the mask values converge to zero, some mask values may be exactly zero while the others are close to zero. We clip the masks with small values to zero according to the pruning budget to make sure the corresponding channels are completely removed. Then, we optimize the sub-problem for $\widehat{\bm{W}}_k^{(i)}$ until converge. The final weights of the pruned layer are obtained by applying the mask $\hat{\bm\beta}_k^{(i)}$ to the weights $\widehat{\bm{W}}_k^{(i)}$.

\section{Experiments}

\begin{table}[t!]
    \centering
    \caption{Dataset statistics. The Attr. column shows the dimension of the node attributes. (s) in Classes denotes multi-class single-label classification problem while (m) denotes multi-class multi-label classification problem. The Test\% column shows the percentage of test nodes.}
    \begin{tabular}{r|c@{\hspace{1.2ex}}c@{\hspace{1.2ex}}c@{\hspace{1.2ex}}c@{\hspace{1.2ex}}c}
        Dataset & Nodes & Edges & Attr. & Classes & Test\%\\
        \toprule
        Flickr & 89,250 & 899,756 & 500 & 7(s) & 25\%\\
        Arxiv & 169,343 & 1,166,243 & 128 & 40(s) & 29\%\\
        Reddit & 232,965 & 11,606,919 & 602 & 41(s) & 24\%\\
        Yelp & 716,847 & 6,977,410 & 300 & 100(m) & 10\%\\
        Products & 2,449,029 & 61,859,140 & 100 & 47(s) & 88 \%\\
        \midrule
        \textcolor{reb}{YelpCHI} & \textcolor{reb}{67,395} & \textcolor{reb}{287,619} & \textcolor{reb}{769} & \textcolor{reb}{2(s)} & \textcolor{reb}{23\%}\\
    \end{tabular}
    \label{tab: dataset}
\end{table}

\begin{table*}
    \setlength{\tabcolsep}{1.35mm}
    \fontsize{9}{9}\selectfont
    \centering
    \caption{\textcolor{reb}{Pruned full inference results on GPU. The $\ast$ nodes in the plots denote the results of the reference models (no pruning).}}
    \begin{tabular}{c|cccc|cccc|cccc|cccc}
        & & \multicolumn{2}{c}{Flickr} & & & \multicolumn{2}{c}{Arxiv} & & & \multicolumn{2}{c}{Reddit} & & & \multicolumn{2}{c}{Yelp}\\
        \toprule
        Budget & - & 2$\times$ & 4$\times$ & 8$\times$ & - & 2$\times$ & 4$\times$ & 8$\times$ & - & 2$\times$ & 4$\times$ & 8$\times$ & - & 2$\times$ & 4$\times$ & 8$\times$\\
        \midrule
        \multirow{1}{*}{F1-Micro} & 0.511 & \textcolor{reb}{0.517} & \textcolor{reb}{0.520} & \textcolor{reb}{0.517} & 0.716 & 0.712 & \textcolor{reb}{0.710} & \textcolor{reb}{0.706}   & 0.966 & 0.966 & \textcolor{reb}{0.964} & \textcolor{reb}{0.959} & 0.654 & \textcolor{reb}{0.654} & \textcolor{reb}{0.651} & \textcolor{reb}{0.640}\\
        \midrule
        \#kMACs/node & 545 & 211 & 94 & 48 & 1242 & 360 & 115 & 40 & 317 & 172 & 112 & 85 & 1490 & 485 & 180 & 77\\
        \midrule
        Mem. (MB) & 531 & 269 & 221 & 199 & 1997 & 1002 & 505 & 257 & 852 & 738 & 681 & 652 & 8459 & 4256 & 2155 & 1225\\
        \midrule
        Thpt. (mN/s) & 2.69 & 5.28 & 9.11 & 15.13 & 1.11 & 1.98 & 3.79 & 6.72 & 2.47 & 4.30 & 7.17 & 10.35 & 0.90 & 1.95 & 3.05 & 4.40\\
        \midrule
        Thpt. Impr. & - & 1.96$\times$ & 3.39$\times$ & 5.63$\times$ & - & 1.79$\times$ & 3.42$\times$ & 6.07$\times$ & - & 1.74$\times$ & 2.90$\times$ & 4.19$\times$ & - & 2.16$\times$ & 3.38$\times$ & 4.82$\times$ \\
        \midrule
        \begin{tabular}{c}F1mic-Thpt\end{tabular} & 
        \multicolumn{4}{c|}{
            \begin{tabular}{c}
                \includegraphics{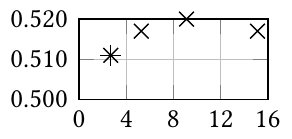}
            \end{tabular}
        } & 
        \multicolumn{4}{c|}{
            \begin{tabular}{c}
                \includegraphics{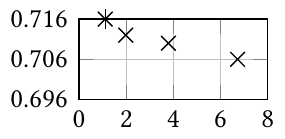}
            \end{tabular}
        } & 
        \multicolumn{4}{c|}{
            \begin{tabular}{c}
                \includegraphics{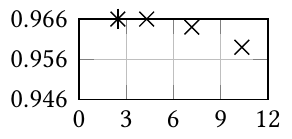}
            \end{tabular}
        } & 
        \multicolumn{4}{c}{
            \begin{tabular}{c}
                \includegraphics{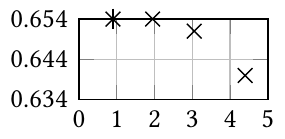}
            \end{tabular}
        }
    \end{tabular}
    \label{tab: full}
\end{table*}

We evaluate the performance of the proposed method with the node classification problem on five popular datasets: 1. \textbf{Flickr} \cite{graphsaint} classifying the types of user-uploaded images, 2. \textbf{Arxiv} \cite{hu2020open} classifying subject areas of Arxiv CS papers, 3. \textbf{Reddit} \cite{graphsage} classifying communities of Reddit posts, 4. \textbf{Yelp} \cite{graphsaint} classifying types of businesses on Yelp, 5. \textbf{Products} \cite{hu2020open} classifying categories of products on Amazon. \textcolor{reb}{For batched inference, we also evaluate with a real world spam detection application on the YelpCHI\cite{yelpchi} dataset that identifies spam reviews on Yelp.} We adopt supervised and inductive settings on all datasets.

For the models to prune, we use the 2-layer GraphSAGE \cite{graphsage} architecture (Equation \ref{eq: forward} with $K'=0,K=1$) with the common hidden feature size $256,512,128,512,512$ on the five \textcolor{reb}{node classification} datasets, respectively. 
\textcolor{reb}{On the YelpCHI dataset, we use $128$ as the hidden feature size.}
We use the standard single floating point precision for both the original models and the pruned models. To obtain trained models to prune, we adopt the sub-graph based training technique from GraphSAINT \cite{graphsaint} with the random walk sampler. For each dataset, we prune with three global budgets $\eta=0.5,0.25,0.125$ and obtain three pruned models ($2\times,4\times,8\times$) in different sizes. We choose 1024 as the batch size and use the ADAM optimizer for SGD in the two sub-problems. After pruning, we re-train the pruned models until convergence. 

To test the speedup of the pruned models, we measure the throughput and latency of full inference on the first four datasets with GPU, and batched inference on all five datasets with CPU and GPU. For batched inference, we form batches randomly from the nodes in the test set until all the nodes in the test set are covered. All accuracy (F1-Micro) results are for the test nodes only. The pruning framework is implemented using PyTorch and Python3. \textcolor{reb}{We run all experiments on a machine with 64-core ThreadRipper 2990WX CPU with 256GB of DDR4 RAM, and a single NVIDIA RTX A6000 GPU with 48GB of GDDR6 RAM.} All the accuracy results are averages of three runs. For batched inference, we limit the number of hop-2 neighbors to be 32.

\subsection{Performance of Single Layer Pruning}

\begin{figure}[b]
\pgfplotsset{compat=1.5}

\includegraphics{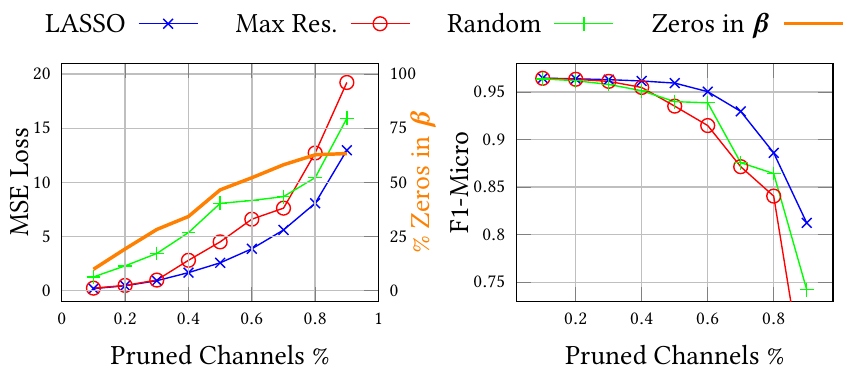}
    \caption{Loss and F1-Micro curves under different numbers of pruned channels in layer-$2$ on the Reddit dataset. \textcolor{reb}{We also show the percentage of $\bm\beta$ that shrinks to zero for the LASSO pruning method in the left figure.}}
    \label{fig: loss-acc}
\end{figure}

We compare the proposed pruning method (LASSO) with pruning the channels with small L1-norm in the corresponding weight matrix (Max Res.) and randomly pruning the channels (Random). Figure \ref{fig: loss-acc} shows the loss and F1-Micro curves under different numbers of pruned channels in both branches of layer-$2$ on the Reddit dataset. We apply layer-wise re-training for all three pruning methods. The proposed pruning method clearly outperforms other pruning methods, especially when the number of pruned channels is more than $30\%$.

\subsection{Full Inference}

\begin{table*}
    \setlength{\tabcolsep}{1.1mm}
    \centering
    \caption{\textcolor{reb}{Pruned batched inferences results on CPU (batch size=512). The second rows in each metric show the results with stored hidden features. The $\ast$ nodes in the plots denote the results of the reference models (no pruning, w/o store).}}
    \fontsize{9.3}{9.3}\selectfont
    \begin{tabular}{c|cccc|cccc|cccc|cccc}
        & & \multicolumn{2}{c}{Arxiv} & & & \multicolumn{2}{c}{Reddit} & & & \multicolumn{2}{c}{Yelp} & & & \multicolumn{2}{c}{Products}\\
       \toprule
       Budget & - & 2$\times$ & 4$\times$ & 8$\times$ & - & 2$\times$ & 4$\times$ & 8$\times$ & - & 2$\times$ & 4$\times$ & 8$\times$ & - & 2$\times$ & 4$\times$ & 8$\times$\\
       \midrule
       F1-Micro & 0.714 & 0.710 & 0.709 & 0.707 & 0.966 & 0.966 & 0.964 & 0.955 & 0.654 & 0.654 & 0.652 & 0.646 & 0.792 & 0.791 & 0.785 & \textcolor{reb}{0.764}\\
       w/ store & 0.714 & 0.710 & 0.709 & 0.707 & 0.966 & 0.966 & 0.964 & 0.954 & 0.653 & 0.653 & 0.652 & 0.646 & 0.792 & 0.792 & 0.786 & \textcolor{reb}{0.766}\\
       \midrule
       \#kMACs/node & 3135 & 1620 & 846 & 395 & 17665 & 7409 & 3288 & 1052 & 7870 & 3696 & 1650 & 840 & 3952 & 2044 & 1090 & 520\\
       w/ store & 2118 & 1096 & 577 & 286 & 6225 & 2627 & 1171 & 381 & 3908 & 1888 & 895 & 485 & 1590 & 827 & 446 & 240\\
       \midrule
       Mem. (MB) & 85 & 49 & 30 & 14 & 3086 & 1551 & 790 & 409 & 225 & 122 & 53 & 26 & 96 & 65 & 49 & 28\\
       w/store & 72 & 42 & 23 & 10 & 1431 & 568 & 288 & 147 & 165 & 92 & 39 & 19 & 70 & 37 & 21 & 10\\
       \midrule
       Lat. (ms) & 27 & 20 & 17 & 13 & 411 & 217 & 128 & 85 & 56 & 38 & 25 & 16 & 120 & 58 & 48 & 35\\
       w/ store & 15 & 8 & 6 & 5 & 101 & 56 & 34 & 24 & 22 & 14 & 10 & 7 & 47 & 30 & 27 & 20\\
       \midrule
       Lat. Impr. & - & 1.33$\times$ & 1.54$\times$ & 2.12$\times$ & - & 1.90$\times$ & 3.22$\times$ & 4.86$\times$ & - & 1.46$\times$ & 2.20$\times$ & 3.59$\times$ & - & 2.09$\times$ & 2.50$\times$ & 3.42$\times$\\
       w/ store & 1.82$\times$ & 3.24$\times$ & 4.18$\times$ & 5.29$\times$ & 4.09$\times$ & 7.35$\times$ & 12.26$\times$ & 17.04$\times$ & 2.50$\times$ & 3.84$\times$ & 5.84$\times$ & 8.08$\times$ & 2.56$\times$ & 3.99$\times$ & 4.40$\times$ & 6.02$\times$ \\
       \midrule
       \begin{tabular}{c}F1mic-Lat.\\\textcolor{blue}{w/o store}\\\textcolor{red}{w/ store}\end{tabular} & 
        \multicolumn{4}{c|}{
            \begin{tabular}{c}
                \includegraphics{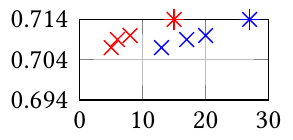}
            \end{tabular}
        } & 
        \multicolumn{4}{c|}{
            \begin{tabular}{c}
                \includegraphics{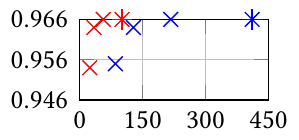}
            \end{tabular}
        }  & 
        \multicolumn{4}{c|}{
            \begin{tabular}{c}
                \includegraphics{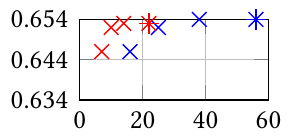}
            \end{tabular}
        } & 
        \multicolumn{4}{c}{
            \begin{tabular}{c}
                \includegraphics{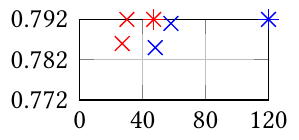}
            \end{tabular}
        }
    \end{tabular}
    \label{tab: batched}
\end{table*}

Table \ref{tab: full} shows the results for full inference on GPU. The computation complexity is measured in thousands of Multiplication-and-ACcumulation operations per node (\#kMACs/node). The throughput is measured in thousand of target nodes computed per second (kN/s) or million of target nodes computed per second (mN/s). For memory usage, we adopt in-place point-wise operations without storing the intermediate values as we only need to compute forward propagation at inference. The latency is the GPU execution time of a complete forward propagation. The throughput and memory usage are calculated for all the nodes in the graphs. 
In the F1mic-Thpt. row, the x and y axes are the throughput (in mN/s) and F1-micro.
On the Flickr dataset, the pruned models achieve higher F1-Micro than the original models, possibly due to better convergence of smaller models. 
The reduction in computation and memory usage depends on the dimension of the input node attributes. 
We achieve close to $\eta^2$ reduction in computation complexity and $\eta$ reduction in memory usage on the Arxiv and Yelp dataset with small input node attributes dimensions. 
We achieve an average of $3.27\times$ speedup on GPU with less than 0.006 drop in F1-Micro for all datasets with $4\times$ pruned models. On the Flickr, Arxiv and Reddit datasets, the $8\times$ pruned models still achieve similar accuracy as the original models.
The pruned models for full inference reduce the latency on small datasets to meet the requirements for real-time applications and increase the throughput for large datasets.
The pruned models also make it possible to run full batch inference of small graphs on edge devices with limited memory. 
\textcolor{reb}{We observe consistent GPU utilization of around 50\% for models with different pruning budgets.}

\textcolor{reb}
{
    On the Flickr, Arxiv, Reddit, and Yelp dataset, our pruning method takes 2.35, 4.34, 6.35, and 32.15 seconds in pruning and 1.36, 6.38, 10.02, and 346.21 seconds in re-training.
    Due to the small number of parameters in the pruned models, the re-training of the pruned models takes less time than training the original models.
}

\subsection{Batched Inference}
\label{sec: batchedinf}

Table \ref{tab: batched} shows the results for batched inference on CPU. We calculate the memory usage by the amount of memory needed to compute the forward path of one batch. The attributes and stored hidden features of the supporting nodes in each batch are fed into CPU from DDR4 (peak bandwidth 68GB/s) and GPU from GDDR6 (peak bandwidth 768GB/s) memory.
In the F1mic-Thpt. row, the x and y axes are the throughput (in kN/s) and F1-micro.
We achieve $\eta$ reduction in computation complexity and memory usage on all datasets for batched inference without stored hidden features. We store the hidden features of training and validation nodes, and the root nodes in each batch of inference. The storing of hidden features further reduces an average of $33\%$ of supporting nodes in layer-$1$.
We reduce the memory usage from 85-3086MB to 10-147MB, which makes it possible to perform inference on edge devices like mobiles. The memory usage also reflects an upper bound of the amount of input node attributes needed to perform one batch of forward propagation. 
On all five datasets, the pruned models with stored hidden features achieve less than 30ms (up to $17\times$ improvement) latency on CPU with less than 0.012 F1-Micro drop. The pruned models with stored hidden feature meet the requirements of most real-time applications on CPU.
\textcolor{reb}{On GPU, our $4\times$ models achieve $4,16,4,8$ms latency without stored hidden features and $4,6,3,6$ms latency with stored hidden features on the Arxiv, Reddit, Yelp, and Products datasets, respectively.}
Compared with full inference, batched inference requires less memory and computation for a small number of target nodes. For latency-sensitive or large scale applications, batched inference provides a light-weight and low-latency solution.
\textcolor{reb}{We observe 100\% CPU utilization on all models, and 20\% to 50\% GPU utilization on GPU depending on the model size.}

\begin{figure}[b]
\pgfplotsset{compat=1.5}

\includegraphics{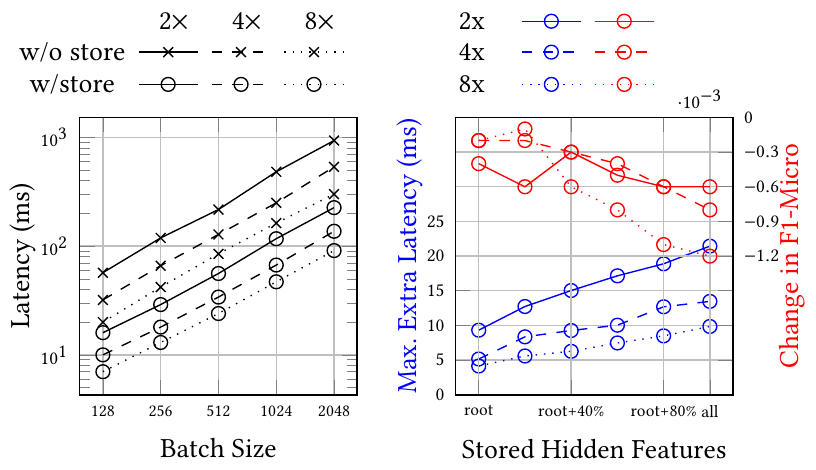}
    \caption{\textcolor{reb}{(a). Latency curves under different batch sizes on the Reddit dataset on CPU. (b). Maximum extra latency and accuracy drop curves with different percentages of stored hidden features.}}
    \label{fig: batchsize}
\end{figure}

Figure \ref{fig: batchsize}.a shows the latency under different batch sizes on the Reddit dataset on CPU. 
\textcolor{reb}{The latency grows linearly with the batch size, which shows that our pruning method accommodates applications with different inference batch sizes.}
\textcolor{reb}{Figure \ref{fig: batchsize}.b shows the trade-off between storing hidden features and extra latency and drop in F1-Micro. Note that the extra latency is mostly caused by the storing of the hidden features, which can be done offline. }

\subsubsection{Spam Detection Application}
\label{sec: yelpchi}

\begin{figure}[t]
    \centering
\pgfplotsset{compat=1.5}

\includegraphics{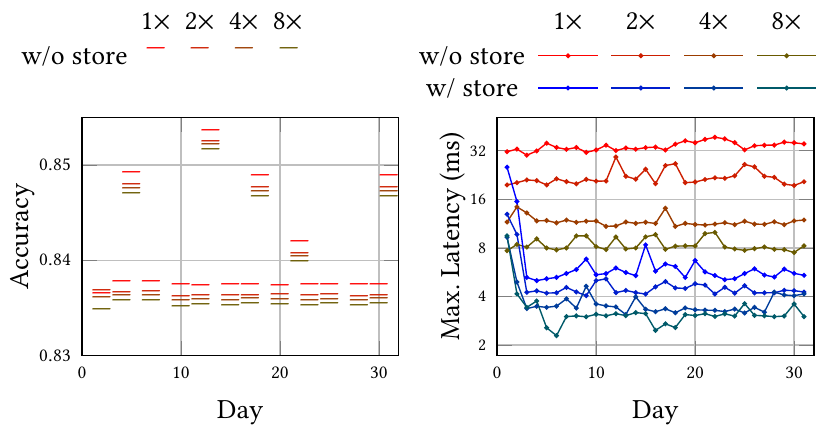}
    \caption{\textcolor{reb}{Accuracy and maximum latency of each day in the first month on the YelpCHI dataset. We only show the accuracy without stored hidden features because the accuracy with stored hidden features is very close. }}
    \label{fig: yelpchi}
\end{figure}

\textcolor{reb}{To evaluate the performance of the proposed pruning technique on real-time applications, we over-sample the YelpCHI \cite{yelpchi} dataset 400 times to create a graph with 27 million nodes which has similar scale as the Yelp website. The nodes in the graph represent reviews for restaurants and hotels in Chicago and are attached with timestamps. The task is to identify spam reviews from the posted reviews between October 2011 and October 2012. We adopt the strategy to perform inference on the emerging reviews every 30 minutes and re-train the model every month. The $1\times$(reference), $2\times$, $4\times$, and $8\times$ models achieve 0.873, 0.871, 0.866, and 0.865 accuracy on the test set. Figure \ref{fig: yelpchi} shows the accuracy and maximum latency of each day in the first month. For inference with stored hidden features, the first few days have higher latency due to the indexing and storing of the hidden features. However, the latency is still lower than inference without stored hidden features, even in the first few days.}

\subsection{Comparison with Other GNNs}

We compare the throughput and accuracy of full inference with GAT \cite{gat}, SIGN \cite{frasca2020sign}, Jumping Knowledge Network (JK) \cite{jumpingknowledge}, GraphSAGE \cite{graphsage}, PPRGo \cite{PPRGo} \footnote{We tune the parameters of PPRGo to fit the supervised learning setting.}, GCN \cite{kipfgcn}, SGC \cite{sgc}, and \textcolor{reb}{TinyGNN \cite{tinygnn}}. We use a similar two-layer (or equivalent of two-hop propagation) architecture on all baselines \textcolor{reb}{except a one-layer student network supervised by a two-layer teacher network for TinyGNN}. Figure \ref{fig: acct} shows the inference throughput and accuracy of various GNN architectures on the Reddit dataset on GPU. Our $4\times$ model achieves top-tier accuracy, comparable with GAT, SIGN and GraphSAGE, but with significant improvement in throughput ($6.96\times,4.74\times,2.59\times$ with GAT, SIGN, and GraphSAGE, respectively). 

\newcommand{\mcrot}[4]{\multicolumn{#1}{#2}{\rlap{\rotatebox{#3}{#4}~}}} 
\begin{table}[t]
    \centering
    \begin{tabular}{c|c@{\hspace{-1ex}}c@{\hspace{2ex}}c@{\hspace{1ex}}c}
        & & Pre-Proc. & F1-micro & \#kMACs/node\\
        \toprule
        \multirow{5}{*}{\rotatebox[origin=c]{90}{\parbox{1cm}{Full Inf.}}} & \multirow{2}{*}{SGC} & - & \multirow{2}{*}{0.949} & 146 \\
        & & \cmark & & 25 \\
        \cline{2-5}
        & \multirow{2}{*}{SIGN(2,0,0)} & - & \multirow{2}{*}{0.966} & 978 \\
        & & \cmark & & 858 \\
        \cline{2-5}
        & PPRGo & - & 0.937 & 148\\
        \cline{2-5}
        & \textcolor{reb}{TinyGNN} & \textcolor{reb}{-} & \textcolor{reb}{0.957} & \textcolor{reb}{273} \\
        \cline{2-5}
        & ours-4$\times$ & - & 0.964 & 112 \\
        \toprule
        \multirow{3}{*}{\rotatebox[origin=c]{90}{\parbox{1cm}{Batched Inf.}}}& MLP-2 & - & 0.702 & 120\\
        \cline{2-5}
        & ours-4$\times$ w/o& - & 0.964 & 3288 \\
        & ours-4$\times$ w/& - & 0.964 & 1171 \\
    \end{tabular}
    \caption{Comparison of accuracy and per node computation for full inference and batched inference on the Reddit dataset. \textcolor{reb}{The w/ and w/o in batched inference denotes with and without stored hidden features.}}
    \label{tab: comp}
\end{table}

\subsubsection{Computation Comparison with Simplified GNNs}
We compare the accuracy and per node computation on the Reddit dataset of our 4$\times$ pruning models with SGC, SIGN with $(r,s,t)=(2,0,0)$, PPRGo with two-pass inference, \textcolor{reb}{TinyGNN with a 1-layer PAM student network supervised by a 2-layer teacher network,} and 2-layer MLP with 128 hidden features (MLP-2). 
\textcolor{reb}{Table \ref{tab: comp} shows the result of the comparison with other simplified GNNs.}
The pre-processing for both SGC and SIGN is to twice compute feature propagation ($\tilde{\bm{A}}^2\cdot\bm{h}^{(0)}$) for 120 kMACs/node. If any graph structure or node attributes change, the pre-processing needs to be re-computed. For SGC, if the input node features are pre-processed, there is only one MLP layer transforming the aggregate features to class probabilities, leading to the lowest computation. SIGN has the highest per node computation as the numbers of hidden units in the feedforward layers are high (460 for GNN layers and 675 for the classification layer). 
For full inference, our pruned model achieves higher accuracy than SGC and TinyGNN, and comparable accuracy to SIGN with less computation. For batched inference, our pruned models achieve remarkably higher accuracy than MLP.

\section{Conclusion}

We presented a novel method of pruning the input channels to accelerate large scale and real-time GNN inference. We formulated the GNN pruning problem as a LASSO optimization problem to select from the input channels to approximate the output. We developed different pruning schemes according to the computation complexity and memory usage in different inference scenarios.
We designed a unique technique for batched inference to further reduce computation by storing and reusing the hidden features. We conducted experiments on real-world datasets to demonstrate that the pruned models greatly reduce computation and memory usage while still maintaining high accuracy. We showed the improvement on latency and throughput of using the pruned models on CPU and GPU. The light-weight pruned models are attractive to energy-efficient devices like mobile processors and FPGA, as well as 
applications like real-time recommendation and fraud detection on social networks. 

\begin{acks}
  This work is supported by the National Science Foundation (NSF) Research Fund of SPX: Collaborative Research: FASTLEAP: FPGA based compact Deep Learning Platform (Number CCF-1919289). Any opinions, findings and conclusions or recommendations expressed in this work are those of the authors and do not necessarily reflect the views of NSF.
\end{acks}

\bibliographystyle{ACM-Reference-Format}
\bibliography{reference}

\end{document}